\documentclass[letterpaper, 10 pt, conference]{ieeeconf}
\IEEEoverridecommandlockouts                              

\usepackage{cite}

\usepackage{amsmath,amssymb,amsfonts}
\usepackage{algorithmic}
\usepackage{graphicx}
\graphicspath{ {./figures/} }
\usepackage{textcomp}
\usepackage{xcolor}
\usepackage{bbm}
\usepackage{comment}
\usepackage{paralist}
\usepackage[caption=false]{subfig}
\usepackage{scalerel}    

\usepackage{amsthm}
\theoremstyle{plain}

\usepackage[linesnumbered,ruled]{algorithm2e}

\theoremstyle{remark}
\newtheorem{remark}{Remark}
\theoremstyle{definition}
\newtheorem{definition}{Definition}
\theoremstyle{plain}

\usepackage{dirtytalk}
\usepackage[hidelinks]{hyperref}
\newcommand{\todo}[1]{{\color{red}{#1}}}

\let\oldnl\nl
\newcommand{\nonl}{\renewcommand{\nl}{\let\nl\oldnl}}

\DeclareMathOperator*{\argmax}{argmax} 
\usepackage{mathtools}
\DeclarePairedDelimiter\abs{\lvert}{\rvert}%
\usepackage{bm}

\DeclarePairedDelimiterX{\norm}[1]{\lVert}{\rVert}{#1}

\usepackage{mathrsfs}
\usepackage[para,online,flushleft]{threeparttable}
\usepackage{array,booktabs,makecell}
\usepackage{diagbox}

\usepackage{soul}

\usepackage{bbm}

\begin{document}

\title{\LARGE \bf 
Decision-Oriented Learning Using Differentiable Submodular Maximization for Multi-Robot Coordination
\thanks{This work is supported in part by National Science Foundation Grant No. 1943368 and Army Grant No. W911NF2120076.}
\thanks{\textsuperscript{1}University of Southern California, Los Angels, CA 90007 USA {\tt\small shig@usc.edu}. The author was with the Department of Electrical and Computer Engineering, University of Maryland, College Park, MD, USA, when part of the work was completed.}
\thanks{\textsuperscript{2}University of Maryland, College Park, MD 20742 USA {[{\tt\small cshek1}, {\tt\small tokekar]@umd.edu}}}
\thanks{\textsuperscript{3}Woods Hole Oceanographic Institute, Woods Hole, MA 02543 USA {\tt\small nare@whoi.edu}
}
}

\author{Guangyao Shi,\textsuperscript{1} Chak Lam Shek,\textsuperscript{2} Nare Karapetyan,\textsuperscript{3} Pratap Tokekar\textsuperscript{2} }


\maketitle

\begin{abstract}
We present a differentiable, decision-oriented learning framework for cost prediction in a class of multi-robot decision-making problems, in which the robots need to trade off the task performance with the  costs of taking actions when they select actions to take. 
Specifically, we consider the cases where the task performance is measured by a known monotone submodular function (e.g.,  coverage, mutual information), and the cost of actions depends on the context (e.g., wind and terrain conditions). We need to learn a function that maps the context to the costs.  
Classically, we treat such a learning problem and the downstream decision-making problem as two decoupled problems, i.e., we first learn to predict the cost function without considering the downstream decision-making problem, and then use the learned function for predicting the cost and using it in the decision-making problem. However, the loss function used in learning a prediction function may \textit{not} be aligned with the downstream decision-making.
Good performance in the isolated prediction phase obtained using the loss that is not relevant to the downstream decision-making problem does not necessarily lead to good decisions in the downstream task. 
To this end, we propose a \emph{decision-oriented learning} framework that incorporates the downstream task performance in the prediction phase via a differentiable optimization layer.  The main computational challenge in such a framework is to make the combinatorial optimization, i.e., non-monotone submodular maximization, differentiable. This function is not naturally differentiable. 
We propose the Differentiable Cost Scaled Greedy algorithm (D-CSG), which is a continuous and differentiable relaxation of CSG. We demonstrate the efficacy of the proposed framework through numerical simulations. The results show that the proposed framework can result in better performance than the traditional two-stage approach when the number of samples is small ($<600$), which is the case for most robotic applications, and has comparable performance when the number of samples is large.
\end{abstract}

\section{Introduction}
\begin{figure}[t]
\centering
\includegraphics[width=2.7in]{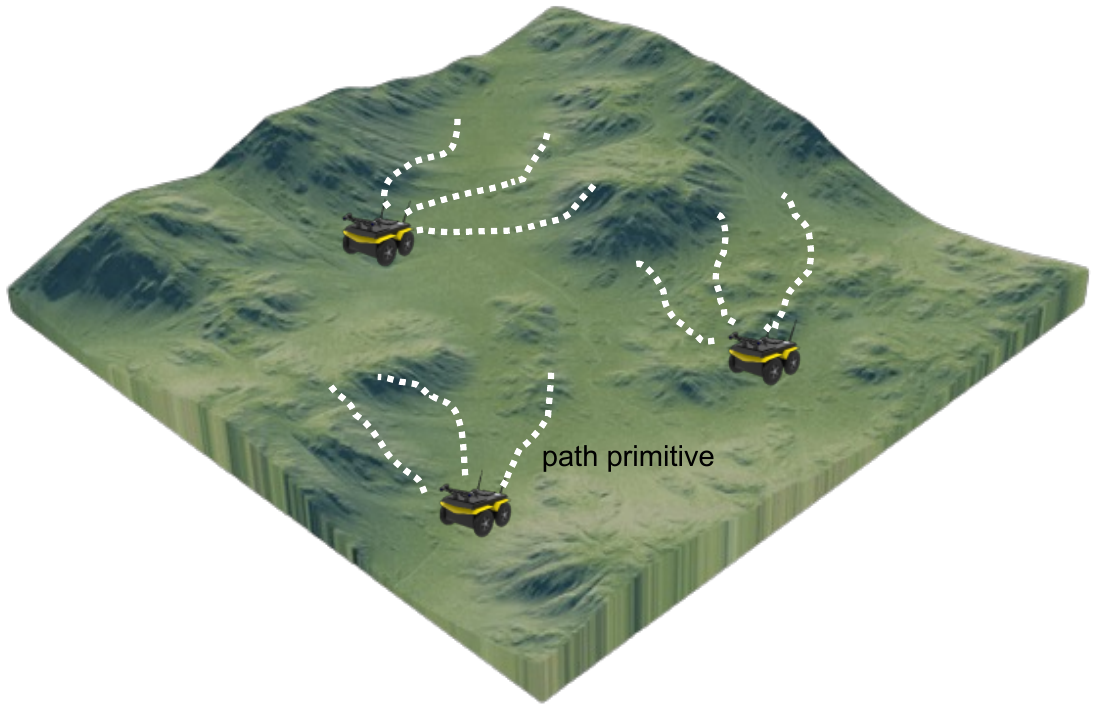}
\caption{The motivating case study of energy-aware multi-robot coordination. A team of robots needs to cover a task area and for every planning epoch, they need to trade off the area they will cover and the energy cost, which depends on the weather and terrain conditions. }
\label{fig:illustration}
\end{figure}

Multi-robot systems are widely used in search and rescue, environmental monitoring, and intelligence surveillance and reconnaissance \cite{zhou2021multi,sung2023survey,queralta2020collaborative, wilde2021learning, shi2022risk}. For these tasks, it is desirable to make robots work as long as possible considering that the robotic platforms have limited energy. To achieve this, when making routing decisions, they should trade off the task performance and the associated energy costs. In this work, we are interested in such cases where the decision-making objective consists of two terms: the first term is a known monotone submodular function, which measures the task performance, and the second term is a linear cost function. If the parameters in the linear cost function are known, an existing algorithm~\cite{nikolakaki2021efficient} can solve the problem with bounded optimality. However, in practice, actual parameters the cost function may be unknown and may depend on the context (e.g., wind, rain, terrain) as shown in Fig. \ref{fig:illustration}. 

Traditionally, we tackle such challenges in two phases: first, we learn a function that maps the context observation to the parameters using historical data without considering the downstream decision-making, and then plug in the learned function into the autonomy pipeline for online decision-making.


\begin{figure}[t]
\centering
\includegraphics[width=3.4in]{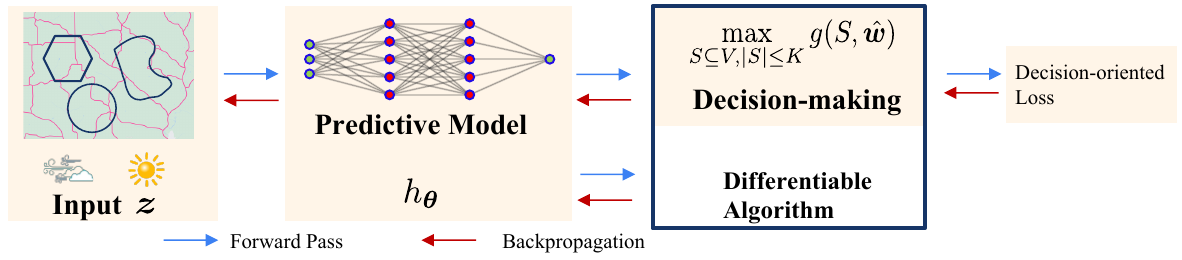}
\caption{The proposed framework that incorporates the non-monotone submodular maximization into the learning process.}
\label{fig:learning_pipeline}
\end{figure}

{Although the two-phase approach has been adopted widely,} recent works have shown that 
incorporating the decision-making task optimization into the learning process for cost prediction can result in better decisions when we use the learned cost prediction function for decision-making ~\cite{wilder2019melding, ferber2020mipaal, mandi2020smart, shi2023decision}. Particularly, the prediction loss function (e.g., Mean Square Error (MSE)) used to learn the cost function may be misaligned with the downstream task. 
In \cite{shi2023decision}, we showed an example where a better predictor trained using MSE results in worse downstream routing decisions. 

An alternative is to solve this problem end-to-end, where we directly map the context input to the decisions using deep neural networks~\cite{mishra2021multi,chen2021multi}. The objective in such an approach is not misaligned since the loss function for training the network will depend directly on the final decisions. However, {the end-to-end approach} faces two limitations. First, training an end-to-end deep neural network to solve a combinatorial optimization problem requires a lot of data~\cite{bengio2021machine}. Secondly, the black-box nature of neural networks makes the resulting decisions hard to explain or interpret. 

Instead, we focus on a \emph{Decision-Oriented Learning (DOL)} framework, as shown in Fig. \ref{fig:learning_pipeline}, that achieves the best of both worlds. We treat the combinatorial optimization problem as a layer in the neural network. By making the optimization differentiable, we train the prediction network using the downstream task-dependent loss rather than the isolated prediction accuracy loss. 
The key challenge here is to devise a differentiable version of an algorithm to solve non-monotone submodular maximization. In our prior work, we showed how to make a monotone submodular function differentiable and use it for multi-robot routing. In this paper, we present a Differentiable Cost-Scaled (D-CSG) algorithm to make the non-monotone maximization differentiable. The differentiability of the non-monotone submodular maximization is achieved by using the multi-linear extension of the set function along with a novel differentiable algorithm, which expands and approximates the existing non-differentiable algorithm~\cite{nikolakaki2021efficient} as a differentiable computational graph.
Our results demonstrate that this approach leads to better downstream decisions than the traditional two-phase approach.

In sum, our contributions include:
\begin{itemize}
    \item We propose a novel differentiable algorithm (D-CSG) based on the Cost-Scaled Greedy (CSG) algorithm.
    \item Based on the proposed differentiable algorithm, we propose a decision-oriented learning framework for predicting intervention costs.
    \item We demonstrate the effectiveness of our framework through extensive numerical simulation.
\end{itemize}

\section{Related Work}\label{sec:related_work}

\textbf{Decision-Oriented Learning} The key idea of decision-oriented learning is to incorporate decision optimization as differentiable layers within the learning pipelines. An obvious advantage of such a learning paradigm is that it enables end-to-end training. Such paradigms are initially studied for continuous optimization problems \cite{amos2017optnet, agrawal2019differentiable}. Subsequently, researchers start to apply such learning paradigms for control and robotics \cite{muntwiler2022learning, amos2018differentiable, chen2018approximating, bhardwaj2020differentiable}. Furthermore, the concept has been extended to combinatorial problems \cite{wilder2019melding, ferber2020mipaal, mandi2020smart, poganvcic2019differentiation}. Our research draws inspiration from  \cite{wilder2019melding} and \cite{ferber2020mipaal}. However, the approaches proposed in \cite{wilder2019melding} and \cite{ferber2020mipaal} cannot deal with non-monotone submodular maximization, which is the focus of our work.

\textbf{Differentiable Submodular Maximization} Submodular maximization and its various adaptations have found widespread application in multi-robot decision-making scenarios encompassing tasks such as coverage, target tracking, exploration, and information gathering.
These studies all benefit from the greedy algorithm
and its variants that can solve submodular maximization problems
efficiently with a provable performance guarantee.
Since the submodular objective and greedy algorithm are tightly coupled, it is better to consider the influence of the greedy algorithm when we consider learning submodular functions \cite{djolonga2017differentiable}. For the non-monotone submodular objective considered in this paper, the simple greedy algorithm \cite{nemhauser1978analysis} does not have a performance guarantee, and we need to use a variant called the CSG algorithm to maximize the objective. As a result, the differentiable versions of the simple greedy algorithm \cite{djolonga2017differentiable, sakaue2021differentiable} cannot be directly used in our learning framework, and we need to develop our differentiable version of the CSG algorithm. Besides, our D-CSG algorithm is technically different from the existing differentiable greedy algorithm. The approach in \cite{djolonga2017differentiable, sakaue2021differentiable} is based on adding stochastic disturbances to the algorithm and using a gradient estimator. Our algorithm is based on the relaxation of the non-differentiable operation to a differentiable operation and the relaxation of the set function to a continuous counterpart.     

\section{Preliminaries}\label{sec:preliminary}
\subsection{Submodular Set Functions}
The definition of submodular functions is given below \cite{krause2014submodular}.
\begin{definition}[Submodularity]
For a set $\mathcal{V}$, a function $f: \{0, 1\}^{\mathcal{V}} \mapsto \mathbb{R}$ is submodular if and only if for any sets $\mathcal{A} \subseteq \mathcal{V}$ and $\mathcal{A}^{\prime} \subseteq \mathcal{V}$, we have $f(\mathcal{A})+f(\mathcal{A}^{\prime}) \geq f(\mathcal{A} \cup \mathcal{A}^{\prime})+f(\mathcal{A} \cap \mathcal{A}^{\prime})$.
\end{definition}

Let $f:\{0, 1\}^\mathcal{V} \to \mathbb{R}_{\geq 0}$ and $c:\{0, 1\}^\mathcal{V} \to \mathbb{R}_{\geq 0}$ be a normalized monotone submodular function and a non-negative linear function, respectively.  We are interested in a special type of submodular function  $g:\{0, 1\}^\mathcal{V} \to \mathbb{R}$, which is defined as 
\begin{equation}\label{eq:main_objective}
    g(\bm{x}, \bm{w}) =  f(\bm{x})-\lambda c(\bm{x}, \bm{w}),
\end{equation}
where $\bm{x} \in \{0, 1\}^\mathcal{V}$; $\lambda$ is a user-specified parameter for their cost tolerance level; and $c(\bm{x}, \bm{w})=\bm{w}^{T}\bm{x}$ and $\bm{w}$ is cost vector for the set $\mathcal{V}$.

It should be noted that $g$ is still a submodular function, but it can take both positive and negative values and may not be monotone \cite{nikolakaki2021efficient, harshaw2019submodular}. Such a function is suitable to model the scenario where we need to balance the task performance ($f(\bm{x})$) with the cost needed to achieve the performance ($c(\bm{x}, \bm{w})$).

The decision-making is to solve the following problem:
\begin{align}\label{eq:main_problem}
    \max_{\bm{x} \in \{0, 1\}^\mathcal{V}} & g(\bm{x}, \bm{w}) \\
\text{s.t.}& ~\bm{1}^{T}\bm{x} \leq K,
\end{align}
where $K$ is the number of elements that can be selected.  

\section{Problem Formulation}\label{sec:problem_formulation}

Our goal is to learn a function  $h_{\bm{\theta}}: \mathcal{Z} \to \mathbb{R}_{+}^{N}$ that maps the context observation $\bm{z} \in \mathcal{Z}$ to the objective parameters $\bm{w} \in \mathbb{R}_{+}^{N}$. Traditionally, finding the mapping $h_{\bm{\theta}}$ and optimizing the downstream objective $g(S, \bm{w})$ are considered separately: (1) given the training data 
$\mathcal{D}=\{(\bm{z}_1, \bm{w}_1), (\bm{z}_2, \bm{w}_2), \ldots, (\bm{z}_{\abs{\mathcal{D}}}, \bm{w}_{\abs{\mathcal{D}}})\}$, find the mapping  $h_{\bm{\theta}}$ by optimizing over $\bm{\theta}$ in a supervised fashion. (2) After optimization, use the parameter $\bm{w}=h_{\theta}(\bm{z})$ for decision-making using (solve Eq. \eqref{eq:main_problem}) when we get an observation $\bm{z}$. 

However, in robotic applications, the available training data is usually limited. Such a pipeline may result in a $h_{\bm{\theta}}$ that either overfits the data or cannot generalize well when deployed, i.e., leads to low-quality decisions in the downstream task. At a high level, the question that we will explore in this paper is:
\begin{center}
    \textit{Can we improve the decision quality in the downstream tasks if we explicitly incorporate the downstream optimization into the process of learning  $h_{\bm{\theta}}$}?
\end{center}

Our answer is that optimizing the following decision-oriented loss can improve the decision quality compared to the baseline approach.

\noindent \textbf{Decision-Oriented Loss:} For given training data $(\bm{z}_i, \bm{w}_i)$, the decision-oriented loss
$\ell_{DOL}(\bm{w}_i, \hat{\bm{w}}_i)$ is defined through Eq. \eqref{eq:mapping_theta} to Eq. \eqref{eq:learning_cost}:
\begin{align}
\hat{\bm{w}}_i & \coloneqq h_{\bm{\theta}}(\bm{z}_i) \label{eq:mapping_theta}\\
\hat{\bm{x}} &\coloneqq \bm{x}^*(\hat{\bm{w}}_i) ~~\text{solving \eqref{eq:main_problem}}~ \text{with}~ \bm{w}=\hat{\bm{w}}_i \label{eq:optimality_definition}\\
\ell_{DOL}(\hat{\bm{w}}_i, \bm{w}_i) &\coloneqq g({\bm{x}^*(\bm{w}_i)}, {\bm{w}_i}) -g(\hat{\bm{x}}, {\bm{w}_i}), \label{eq:learning_cost}
\end{align}
 where $\bm{x}^*(\bm{w}_i)$ denotes the solution of \eqref{eq:main_problem} returned by some approximation algorithms with $\bm{w}={\bm{w}}_i$; $g({\bm{x}^*(\bm{w}_i)}, {\bm{w}_i})$ denotes the decision quality when we use the ground truth parameter $\bm{w}_i$ for decisions; $g(\hat{\bm{x}}, {\bm{w}_i})$ denotes the decision quality when we use the predicted parameter $\hat{\bm{w}}_i$ for decisions, i.e., use $\hat{\bm{w}}_i$ to obtain the decision $\hat{\bm{x}}$, but the decision is evaluated w.r.t. the true parameter $\bm{w}_i$.

 The intuition for Eq. \eqref{eq:learning_cost} is that we want to minimize the gap between the decision quality of the true parameters and that of the predicted parameters. One challenge is when we use the chain rule to compute the gradient of the loss function we need to differentiate through the optimization problem (the first term on the r.h.s. of Eq. \eqref{eq:chain_rule_challenge}) as shown in the illustrative computational graph in Fig. \ref{fig:learning_pipeline}.
 \begin{align}
    \frac{\partial \ell_{DOL}}{\partial \bm{\theta}} 
    = \frac{\partial \ell_{DOL}}{\partial \hat{\bm{w}}_i} \cdot \frac{\partial \hat{\bm{w}}_i}{ \partial \bm{\theta}} \label{eq:chain_rule_challenge}
\end{align}
 In the following sections, we will show how to approximately compute the first term on the r.h.s. of Eq. \eqref{eq:chain_rule_challenge}.

\begin{figure}[!t]
 \centering
    \includegraphics[width=0.40 \textwidth]{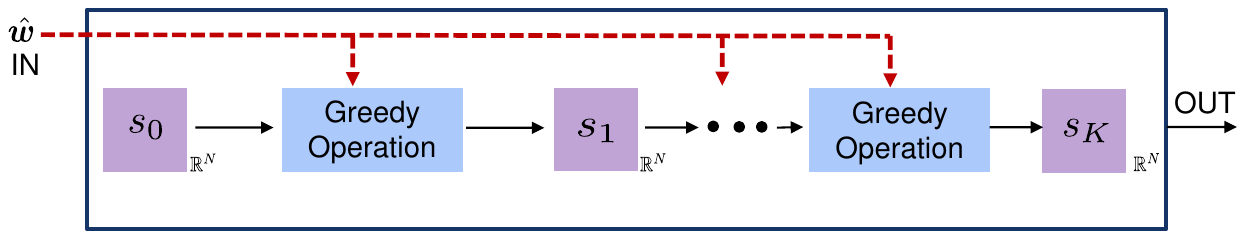}
    \label{fig:diff_CSG}
    \caption{Computational graph of the proposed differentiable algorithm. (a) The structure of the algorithm.}
\end{figure}

\begin{figure}[!t]
 \centering
    
    \includegraphics[width=0.45 \textwidth]{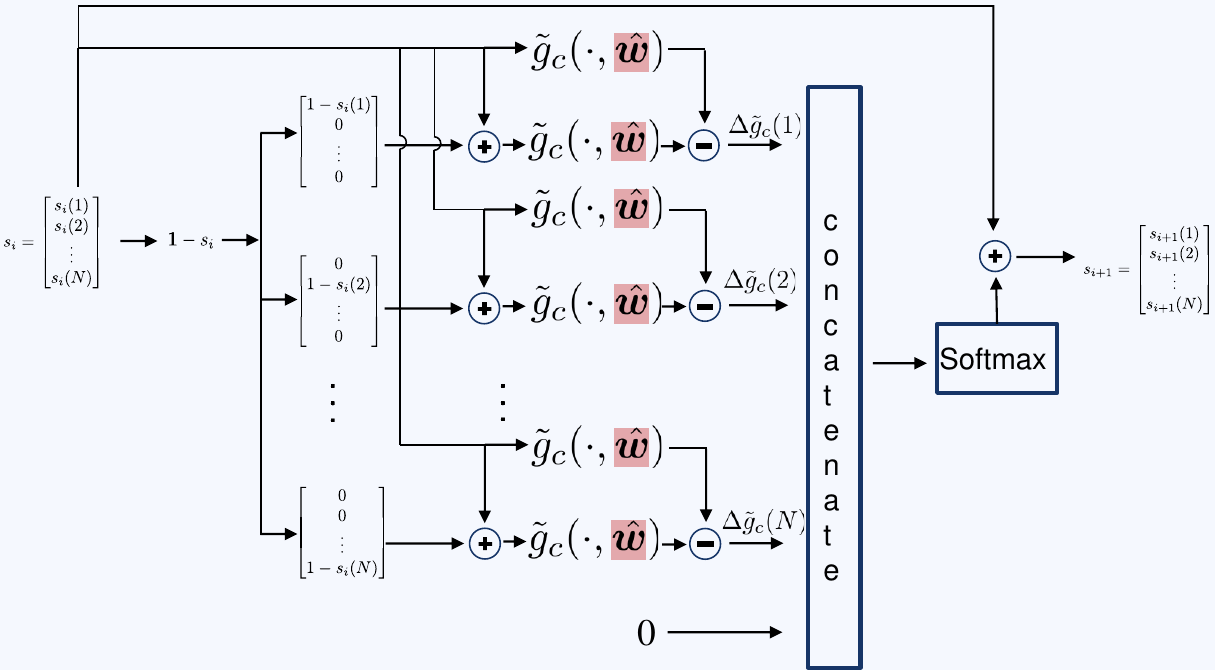}
    \label{fig:diff_greedy}
    
    \caption{Computational graph of the proposed differentiable algorithm. The internal structure of the differentiable cost-scaled greedy operation.}
\end{figure}

\section{Learning Algorithm}\label{sec:learning_algorithm}
We first review the non-differentiable CSG algorithm \cite{nikolakaki2021efficient} for solving non-monotone submodular maximization and then explain the differentiable algorithm used in our decision-oriented learning framework.

\subsection{Cost-Scaled Greedy (CSG) Algorithm}
 \begin{algorithm}[ht]\label{algorithm:scaled_greedy}
    \caption{Cost-Scaled Greedy (CSG) \cite{nikolakaki2021efficient}
    }
    \SetKwInOut{Input}{Input}
    \SetKwInOut{Output}{Output}
    \SetKwProg{Fn}{Function}{:}{}
    \Input{
    Ground set $V$, scaled objective $\Tilde{g}(S, \bm{w})=f(S)-2c(S, \bm{w})$, cardinality $K$
    }
    \Output{A set $S \subseteq V$}
    $S \gets \emptyset$ \\
    \For{i=\rm{1} to \rm{$K$}}{
    $e_i \gets \argmax_{e \in V} \Tilde{g}(e \mid S)$ \\
    \If{$\Tilde{g}(e_i \mid S) \leq 0$}{
    \rm{break}
    }
    $S \gets S \cup \{e_i\}$
    }
    return $S$
\end{algorithm} 
The classic greedy algorithm cannot provide a performance guarantee for the objective in Eq. \eqref{eq:main_objective}. Instead, a modified version of the greedy algorithm, CSG, was proposed in \cite{nikolakaki2021efficient} and was shown to achieve an approximation satisfying that $ f(Q)-c(Q, \bm{w}) \geq \frac{1}{2} f(OPT) -c(OPT, \bm{w})$,
where $Q$ is the solution returned by Algorithm \ref{algorithm:scaled_greedy} and $OPT$ refers to the optimal solution.  
It should be noted that the output of such an algorithm is not differentiable w.r.t. the parameter $\bm{w}$.

\subsection{Multilinear Extension of Submodular Function}
A prerequisite for D-CSG to work is that we need to have a continuous and differentiable relaxation of the objective in Eq. \eqref{eq:main_problem}. The linear part, $c(\bm{x}, \bm{w})$, can be directly relaxed to a continuous version. As for the submodular part, $f(\bm{x})$, We use the multilinear extension to relax the submodular part.  

For a submodular function $f:\{0,1\}^{N} \to \mathbb{R}_{\geq 0} $, its multilinear extension $F:[0,1]^{N} \to \mathbb{R}_{\geq 0} $ is defined as 
\begin{equation}
    F(\bm{x}) = \sum_{\mathcal{S} \subseteq \mathcal{T}} f(\mathcal{S}) \prod_{i \in \mathcal{S}}x_i \prod_{i \notin \mathcal{S}} (1-x_i),
\end{equation}
which is a unique multilinear function agreeing with $f$ in the vertices of the hypercude $[0, 1]^{N}$.

Let $\bm{q}$ denote a random vector in $\{0, 1\}^N$, where each coordinate is independently rouned to $1$ with probability $x_i$ or 0 otherwise. It can be shown that the derivative $\frac{\partial F}{\partial x_i}$ is 
\begin{equation}\label{eq:multilinear_gradient}
    \frac{\partial F}{\partial x_i} = \mathbb{E}_{\bm{q} \sim \bm{x}}\left[  f([\bm{q}]_{i=1})\right]-\mathbb{E}_{\bm{q} \sim \bm{x}}\left[  f([\bm{q}]_{i=0})\right],
\end{equation}
where $[\bm{q}]_{i=1}$ and $[\bm{q}]_{i=0}$ are equal to the vector $\bm{q}$ with the $i$-th coordinate set to be 1 and 0, respectively.

\subsection{Differentiable-Cost-Scaled Greedy (D-CSG) Algorithm}
Based on the CSG algorithm, we develop a differentiable version of CSG. The key idea is to expand the computation steps as one computational graph, as shown in Fig. \ref{fig:diff_CSG}. Suppose we must select up to $K$ elements from a ground set whose size is $N$. We abstract the CSG algorithm as a $K$ step computational graph as shown in Fig. \ref{fig:diff_CSG}. The selection vector is initially an all-zero vector, i.e., $s_0 = \bm{0}, s_0 \in \mathbb{R}^{N}$. In each step, the greedy operation will try to set one element in the selection vector from 0 to 1 approximately. The details of the greedy operation are given in Fig. \ref{fig:diff_greedy}. For an input vector $s_i$, we must first identify the elements that are not selected yet by doing $\bm{1}-s_i$. Then, we separate $\bm{1}-s_i$ into {$N$} vectors, each of which has one element from $\bm{1}-s_i$ and the rest is zero. Each vector represents selecting an element from what is left in the ground set. If an element $s_i(j) \approx 1$ ($j$ is already selected), $1-s_i(j)$ is approximately zero, and the sum of this vector with $s_i$ implies adding no new element to $s_i$. By contrast,  if an element $s_i(j) \approx 0$ ($j$ is not selected yet), $1-s_i(j)$ is approximately one, and the sum of this vector with $s_i$ implies adding one new element to $s_i$. Then, we feed the selection result to the continuous relaxation of the cost-scaled objective function, $\Tilde{g}_c$, to compute the marginal gain. To account for the branch control in Algorithm \ref{algorithm:scaled_greedy} (line 4-6), we add one dummy element with zero marginal gain when we concatenate all the marginal values. Then, this concatenated vector will be fed into one argmax operator to select the one with the largest marginal gain (similar to line 3 in Algorithm \ref{algorithm:scaled_greedy}). If all the marginal gains are less than zero, then the output of the argmax will choose the dummy element. 
As a result,  the first {$N$} elements of the output of the argmax will all approximately to be approximately zero, and the last element corresponding to the dummy selection will be one. Therefore, if we add the result of the first {$N$} elements to $s_i$ to get a new $N$-dimensional vector, $s_{i+1}$, the $s_{i+1}$ will be the same as $s_i$, which is in effect equivalent to skipping selection in this step. Such skipping step will also happen in the following steps since all marginal gains will be less than zero. It is equivalent to the branch control statement in Algorithm \ref{algorithm:scaled_greedy} (lines 4-6). It should be noted that the argmax operator itself is not differentiable and cannot be used during training. Instead, we use Gumbel-softmax \cite{jang2017categorical}, which uses a temperature parameter $\tau$ to scale how it is close to the argmax operator. A larger $\tau$ will make the approximate smoother, but the approximation error will also be larger. In experiments, this parameter is set empirically.  
\begin{remark}
The greedy operation has two non-matrix operations: evaluation of $\Tilde{g}_c$ (2$N$ times) and softmax. The latter is much faster than the former. As a result, the time for evaluation of $\Tilde{g}_c$ will dominate the forward pass of the greedy operation.
\end{remark}

\section{Experiments}\label{sec:experiments}
In this section, we will evaluate the performance of the proposed framework for cost prediction using synthetic data. We will first compare the performances of various algorithms to solve randomly generated instances of Problem \eqref{eq:main_problem} to show the correctness of the D-CSG algorithm. Then, we will present a qualitative example of why the proposed framework is better than the classic one based on MSE loss. Next, we will present some quantitative results to show that the proposed framework leads to better decisions. All experiments were performed on a Windows 64-bit laptop with 16 GB RAM and an 8-core Intel i5-8250U 1.6GHz CPU using Python 3.7.


\begin{figure*}[ht!]
    \centering
    \subfloat[]{
    \includegraphics[width=0.26 \textwidth]{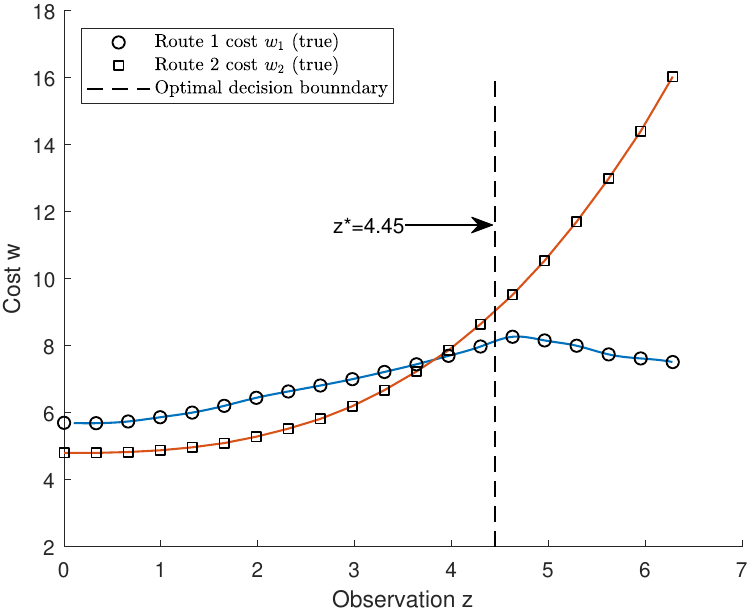}
    \label{fig:SP_ground_truth}
    } 
    \subfloat[]{
    \centering
    \includegraphics[width=0.26 \textwidth]{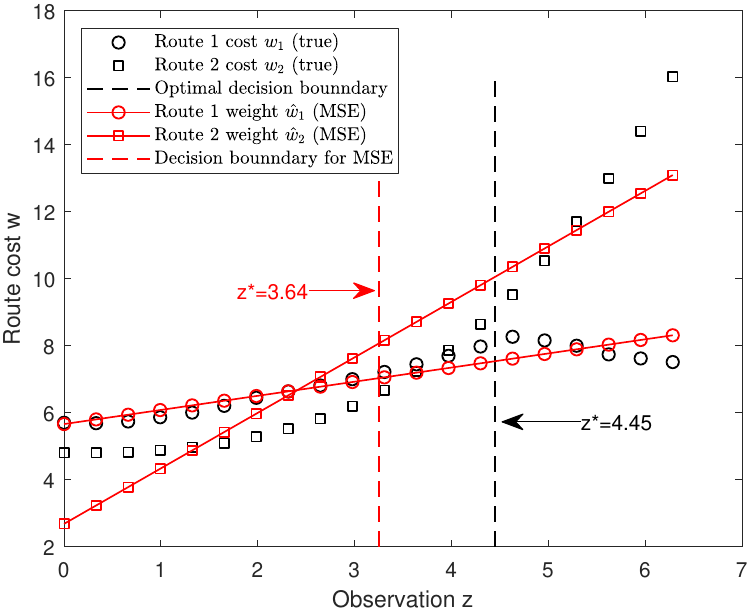}
    \label{fig:MSE_decision}
    }
    \subfloat[]{
    \centering
    \includegraphics[width=0.26 \textwidth]{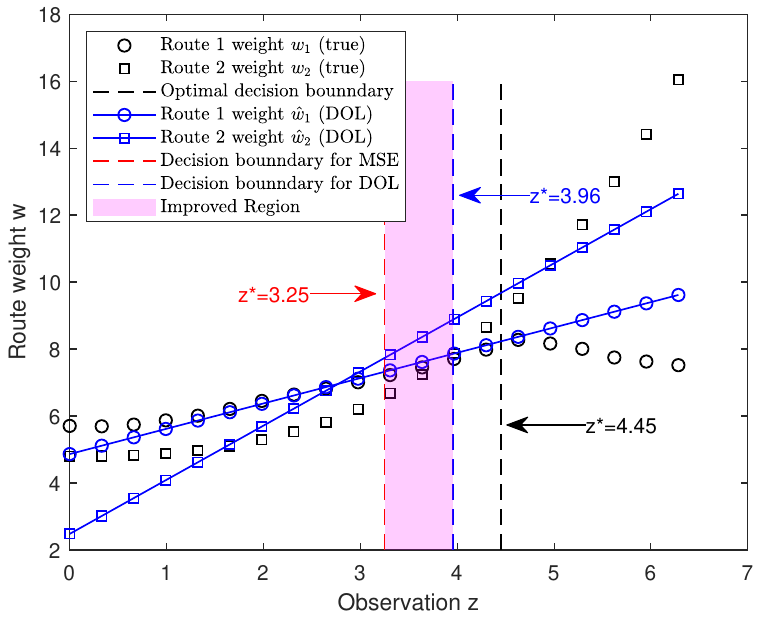}
    \label{fig:LDI_decision}
    }
    \caption{
     A qualitative example to show how DOL is different from the two-stage approach. (a) Ground truth data and the optimal decision boundary.  (b) Learned linear models using MSE loss. (c) Learned linear models using the DOL framework.}
    \label{fig:misalignment_linear_example}
\end{figure*} 

\subsection{Results for D-CSG Algorithm}
\begin{figure}[!t]
\centering
\includegraphics[width=2.1in]{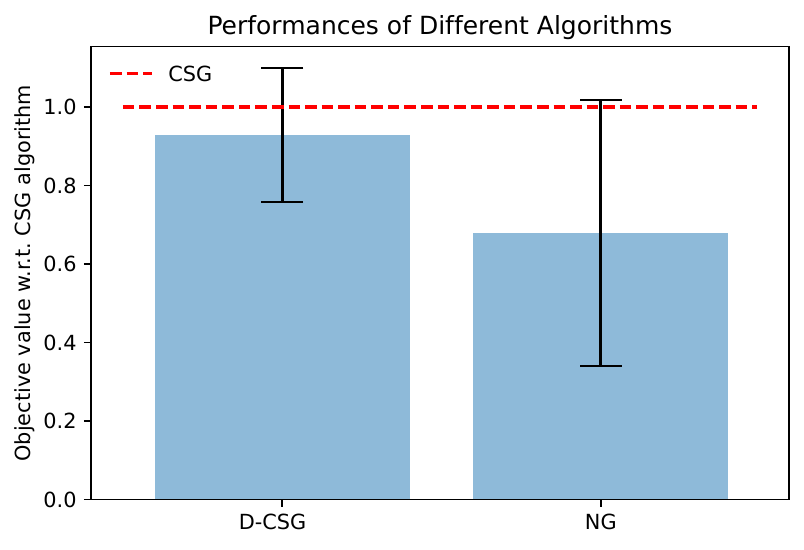}
\caption{Our D-CSG algorithm achieves results closer to the non-differentiable CSG algorithm than the naive greedy (NG).}
\label{fig:algorithm_comparison}
\end{figure}

We test the performance of the proposed differentiable algorithm in synthetic instances of the problem in \eqref{eq:main_problem}. For each instance, we randomly generate a coverage function and associate each element in the ground set a cost. 
We want to answer two questions through experiments. First, since D-CSG is a differentiable approximation of the CSG algorithm, will the differentiability sacrifice much optimality compared to CSG? The second question is what is the running time price for the differentiability.  We compare with two baselines: Naive Greedy \cite{nemhauser1978analysis} and CSG. NG is also non-differentiable; however, there is a differentiable approximation of NG~\cite{sakaue2021differentiable}. By comparing with NG, we essentially are comparing with the upper bound of the differentiable NG algorithm. 

\noindent \textbf{Objective Value} For each instance, we set the objective value returned by CSG as the denominator and scale the outputs of D-CSG and NG. As shown in Fig. \ref{fig:algorithm_comparison}, our D-CSG achieves comparable performance compared to CSG, which suggests that the differentiability does not sacrifice much optimality performance compared to its counterpart CSG. By contrast, the performance of the NG is, on average, worse than that of D-CSG, which justifies our motivation to develop a novel differentiable algorithm rather than using the differentiable version of NG. 

\noindent \textbf{Running Time} The price of differentiability is mainly reflected in the running time. In experiments, we observe that the D-CSG is usually 20-30 times slower than CSG. This is mainly because the evaluation of the continuous relaxation of the submodular objective is time-consuming, which can be viewed as a polynomial with exponentially many terms w.r.t. the size of the ground set. The running time can be improved by using an estimator for function evaluation and gradient computation \cite{ozcan2021submodular}. We also expect this runtime to decrease with better, more optimized code. 
\begin{table}[ht]
\centering
\caption{Submodular Function for the Qualitative Example}
\begin{tabular}[t]{lccccccc}
\toprule
&$s_1$ &$s_2$ &$s_3$ &$s_1, s_2$ &$s_1, s_3$ &$s_2,s_3$ &$s_1, s_2,s_3$\\
\midrule
{$f(\cdot)$} &16 &17 &25 &21  &37  &38 & 41\\
\bottomrule
\end{tabular}
\label{table:qualifative_f}
\end{table}

\subsection{Qualitative Example for DOL Framework}

Let us consider a normalized submodular function $f$, i.e., $f(\emptyset)=0$, defined over a ground set $\mathcal{S}=\{s_1, s_2, s_3\}$. We are interested in solving a problem as defined in Eq. \eqref{eq:main_problem} with $K=2$. Each element in $\mathcal{S}$ can be viewed as as a candidate route for the UGVs out of which two must be chosen. The values of $f$ for choosing different elements are shown in Table \ref{table:qualifative_f}. Verifying the submodularity of $f$ is easy using the definition in Section \ref{sec:preliminary}. Each $\mathcal{S}$ element is associated with a context-dependent cost. 
Suppose that the cost of $s_3$ does not depend on the context and is always equal to one. The ground truth costs for $s_1$ and $s_2$ are context-dependent as shown in Fig.\ref{fig:SP_ground_truth}. 

When we make decisions, we can only see the context, and we need to infer the route costs. The optimal decision is either $\{s_1, s_3\}$ or $\{s_2, s_3\}$ depending on the context (i.e., on the value of z).  If we know the ground truth context-to-cost function as shown in Fig.\ref{fig:SP_ground_truth}, the optimal decision boundary is $z=4.45$ at which the cost choosing $s_2$ is greater than that of $s_1$ by 1. Namely, if the context observation $z$ is less than $4.45$, we should choose $s_2$ and $s_3$. When $z$ exceeds 4.45, we should choose $s_1$ and $s_3$. 

Next, let us look at the result if learning is involved. We want to find a mapping from the observation $z$ to costs. Suppose we obtain the training data by sampling from the ground truth as shown in Fig.\ref{fig:SP_ground_truth}. 
 MSE as the objective for learning without considering the downstream task, we will get two lines as shown in Fig. \ref{fig:MSE_decision}. The decision boundary (dashed vertical red line, $z^*=3.64$, at which $w_2-w_1=1$ ) is on the left of the optimal boundary, thus not optimal. By contrast, if we consider the downstream optimization, we will get two lines as shown in Fig. \ref{fig:misalignment_linear_example} and the decision boundary (dashed vertical blue line, $z^*=3.96$, at which $w_2-w_1=1$ ) is closer to the optimal boundary, thus reducing the regions of suboptimal decisions.

\subsection{Quantitative Results}
In this section, we test the performance of the learned models in a multi-robot coordination problem, in which the coverage function \cite{zhou2018resilient} is chosen as for the task metric. 
We evaluate the models in two aspects: one is the number of samples and the other is the complexity of the model (e.g., the number of layers in a neural network). Specifically, we first generate nonlinear functions $h: \mathcal{Z} \to \mathcal{W}$ as the ground truth function to be learned. Using these functions, we generate datasets $\{(\bm{z}_i, \bm{w}_i)\}$ for training and for test. After training, we test the performance of each model as follows. For each $(\bm{z}_i, \bm{w}_i)$ in the test set, we first compute the solution using CSG and $\bm{w}_i$, denoted as $\mathcal{S}_{CSG}(\bm{w}_i)$. Then we compute the predicted cost parameters $\hat{\bm{w}}_i=h_{\bm{\theta}}(\bm{z}_i)$ using several methods and compute the corresponding solution $\mathcal{S}_{ALG}(\hat{\bm{w}}_i)$. ALG refers to one of four choices: we compare the proposed framework DOL with the two-stage framework. For each we test two neural networks: NN1 consists of one hidden layer (number of neurons: $6\times 40 \times 15$) and NN2 consists of two hidden layers (number of neurons: $6\times 40 \times 40 \times 15$).
The performance of different approaches is measured by
\begin{equation}
    \frac{\abs{g(\mathcal{S}_{CSG}(\bm{w}_i), \bm{w}_i) - g(\mathcal{S}_{ALG}(\hat{\bm{w}}_i), \bm{w}_i)}}{g(\mathcal{S}_{CSG}(\bm{w}_i), \bm{w}_i)}.
\end{equation}
The intuition is we want to measure the normalized deviation from the best result we can obtain if we know the ground truth cost parameters with CSG. 

\begin{figure}[ht!]
    \centering
    \includegraphics[width=0.4 \textwidth]{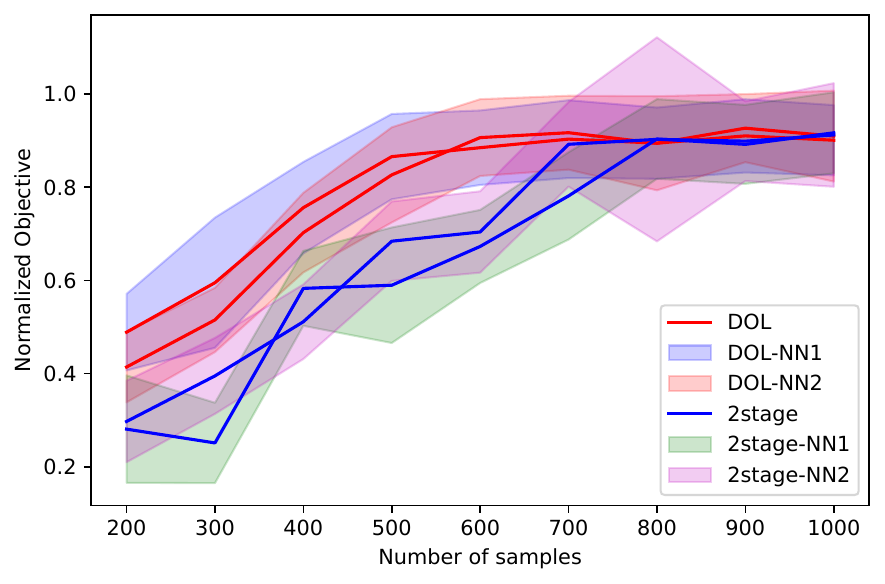}
    
    \caption{
     A case study with three candidate routes and two UGVs. (a) Application scenario. (b) Ground truth data and the optimal decision boundary.  (c) Learned linear models using MSE loss. (d) Learned linear models using the DOL framework.}
    \label{fig:comparison_result}
\end{figure}

As shown in Fig. \ref{fig:comparison_result}, when the sample size is relatively small, which is the usual case for robotic applications, the proposed DOL framework performs better than the two-stage approach. The possible explanation is that incorporating the downstream task into the learning process provides some structural prior and therefore results in better decisions. However, as the number of samples increases, the NN can approximate the function very well and in such cases, different approaches will have similar performance as shown in Fig. \ref{fig:comparison_result}.

\section{Conclusion}
We propose a decision-oriented learning framework for
a special class of routing problems. We first show how to
formulate the learning problem in the context of the vehicle routing problem. Then, we show how to make (non-monotone) submodular maximization a differentiable layer by using the proposed D-CSG algorithm and the multilinear extension of the objective function. The proposed framework and formulation are validated through several case studies.

\bibliographystyle{IEEEtran}
\bibliography{IEEEabrv, main}

\end{document}